\documentclass[runningheads]{llncs}
\usepackage[T1]{fontenc}
\usepackage{graphicx}
\usepackage{booktabs}
\usepackage{amsmath}
\usepackage{algorithm}
\usepackage{algorithmic}
\usepackage{booktabs, tabularx}
\usepackage{subfigure}
\usepackage{multirow, multicol}
\usepackage{amsmath}
\usepackage{colortbl}
\usepackage{bbold}
\usepackage{hyperref}
\usepackage[misc]{ifsym}
\newcommand{\corr}{(\Letter)}
\usepackage{mwe}
\begin{document}

\title{Leveraging Foundation Models for Multi-modal Federated Learning with Incomplete Modality}

\titlerunning{Multi-modal Federated Learning with Incomplete Modality}

\author{Liwei Che\inst{1,2} \and
Jiaqi Wang\inst{2} \and
Xinyue Liu\inst{3} \and
Fenglong Ma\inst{2} \corr
}
\authorrunning{L. Che et al.}

\institute{Rutgers University, Piscataway NJ 08854, USA\\ \email{lw.che@rutgers.edu}
\and
Pennsylvania State University, University Park PA 16802, USA \email{\{jawang,fenglong\}@psu.edu}
\and
Dalian University of Technology, Dalian Liaoning 116621, China
\email{xyliu@dlut.edu.cn}}

\maketitle              % typeset the header of the contribution

\begin{abstract}
Federated learning (FL) has obtained tremendous progress in providing collaborative training solutions for distributed data silos with privacy guarantees. However, few existing works explore a more realistic scenario where the clients hold multiple data modalities. In this paper, we aim to solve a novel challenge in multi-modal federated learning (MFL) -- modality missing -- the clients may lose part of the modalities in their local data sets. To tackle the problems, we propose a novel multi-modal federated learning method, \textbf{Fed}erated \textbf{M}ulti-modal contrasti\textbf{V}e training with \textbf{P}re-trained completion (FedMVP), which integrates the large-scale pre-trained models to enhance the federated training. In the proposed FedMVP framework, each client deploys a large-scale pre-trained model with frozen parameters for modality completion and representation knowledge transfer, enabling efficient and robust local training. On the server side, we utilize generated data to uniformly measure the representation similarity among the uploaded client models and construct a graph perspective to aggregate them according to their importance in the system. We demonstrate that the model achieves superior performance over two real-world image-text classification datasets and is robust to the performance degradation caused by missing modality. 
% \footnote{Code access anonymously at \url{https://anonymous.4open.science/r/FedMVP-6B1C/}}

\end{abstract}

\section{Introduction}

Federated learning (FL) has emerged as a promising paradigm for training machine learning models on decentralized data~\cite{mcmahan2017communication,wang2023knowledge,wang2024rethinking,wang2022towards,wang2024towards,9671374}. In many realistic scenarios, the multi-modal data are collected among distributed data silos and stored in a privacy-sensitive manner, such as the examination and diagnostic records of patients in different hospitals and the multimedia data generated on mobile devices. However, most existing federated learning works focus on single modality scenarios (\textit{e.g.}, image or text) with limited capacity for data with heterogeneous formats and properties. Regarding the fast development of multimedia technology and distributed systems, developing a robust and efficient FL framework for multi-modal machine learning tasks is significant.

To date, several early attempts for multimodal federated learning (MFL) \cite{s23156986} have been proposed \cite{liu2020federated,zong2021fedcmr,yu2023multimodal,xiong2022unified,COBBINAH2022102585,10.1145/3534678.3539384,zhao2022multimodal,DBLP:conf/cikm/Zhou0WH22}. Some of these approaches \cite{10.1145/3534678.3539384,yu2023multimodal,zhao2022multimodal} consider scenarios where the federated system contains both uni-modal and multi-modal clients. However, most of these works assume that all modalities are available to all clients, which is a strong assumption that may not hold in real-world situations. For example, content posted on social media often combines images and text, but users may also publish posts containing only images or text. This modality missing problem poses a substantial challenge as it can severely impact the model's learning ability and performance.

In this paper, we aim to address this general and realistic problem of \textbf{modality missing}, where clients share the same modality combinations, but some multi-modal instances lack part of the modality data. For example, a client holds $1000$ image-text pairs, while $200$ of them only have image data, and $300$ instances have only text data. A few existing works \cite{ma2021smil,Ma_2022_CVPR} focus on the modality incompleteness problem. However, they either only consider text missing in the vision-language learning task or deal with sensor signals that are similar in format. We believe that an advanced MFL framework should be robust to modality incomplete training data and maintain satisfactory performance.

To resolve those challenges, we proposed a multi-modal federated learning framework, namely \textbf{Fed}erated \textbf{M}ulti-modal contrasti\textbf{V}e training with \textbf{P}re-trained completion (FedMVP), which uses frozen pre-trained models as the teachers to support the learnable multi-modal joint encoder module for efficient multi-modal representation learning and to generate informative synthetic data. To enhance the model resilience to the performance degradation caused by modality missing, we utilize the cross-modal generation ability of the recently proposed pre-trained models \cite{radford2021learning,li2022blip,li2023blip} to complete the missing modalities. To further improve the representation learning performance, we proposed an efficient knowledge-transferring method to transfer the representation knowledge from the pre-trained large models to our multi-modal joint learning module. This knowledge-transferring method can alleviate the conflict between the massive data and computing costs requirements for training and fine-tuning of pre-trained large models and the limited resources of federated learning clients. The proposed framework is competent in integrating various pre-trained models with affordable communication costs. As shown in Table~\ref{tab:fp}, compared to the most costly baseline FedViLT, the FedMVP reduces the communication cost by $\boldsymbol{26.7\times}$ and computation FLOPS by $\boldsymbol{15.5\times}$. The pre-trained foundation models will play as the frozen data encoders to transform the original data into high-quality representations, which play an important role in the contrastive-manner training process for the multi-modal joint encoder module. 

\begin{table}[t]
\label{tab:fp}\small
\centering
\caption{Comparison between FedMVP and baselines in terms of \#FLOPS (Floating Point Operations Per Second) and \#transmitted parameters per round. }
\begin{tabular}{l cc}
\toprule
\multicolumn{1}{c}{\textbf{Method}} & \textbf{\#FLOPS} & \textbf{\#Parameters} \\ \midrule
FedViT~\cite{dosovitskiy2020image}  & $22.6G$ & $86.4M$\\
   
FedBERT~\cite{devlin2018bert}   & $38.1G$ & $110.1M$ \\

FedCLIP~\cite{radford2021learning} &  $60.7G$ &$197.2M$\\

FedViLT~\cite{ma2022multimodal} & $55.9G$ &$298.6M$\\

MMFed~\cite{xiong2022unified} & $1.4G$ & $4.49M$\\
\textbf{FedMVP}  & $3.6G$  &$11.2M$\\
     
\bottomrule
\end{tabular}
\end{table}

We summarize our \textbf{contributions} as follows:
(1) We proposed a novel MFL framework that integrates pre-trained large-scale models to conduct efficient multi-modal representation learning and is robust to the modality missing challenge. Our proposed method shows superior performance on two multi-modal classification benchmarks under both IID and non-IID settings. 
(2) To efficiently transfer the learnable representation knowledge from the pre-trained model to the multi-modal joint module under the resource-limited scenario, we proposed a Multi-modal Contrastive Matching (MCM) loss and a Representation Aligned Margin (RAM) loss, which effectively improve the model performance with severe modality missing up to $80\%$.
(3) Instead of aggregating the models based on the data distribution or the model architecture, we propose a novel aggregation algorithm for the MFL server aggregation based on the representation abilities among the client models.

\section{Related Work}

\textbf{Multi-modal Federated Learning (MFL)}.
MFL is still in its early stages of development. Some of the most existing works \cite{xiong2022unified,zong2021fedcmr,liu2020federated} focus on exploring task-specific approaches with complete modalities. In \cite{xiong2022unified}, the authors propose a multi-modal federated learning framework for multi-modal activity recognition with a local co-attention module to fuse multi-modal features. \cite{chen2022towards} gives a detailed analysis of the convergence problem of MFL with late fusion methods under the Non-IID setting. \cite{zhao2022multimodal,10.1145/3534678.3539384,yu2023multimodal} adapt modality-wise encoders to tackle the MFL system with both uni-modal and multi-modal clients. However, few of them explore the scenario where multi-modal data are incomplete, which may cause significant performance degradation.

\noindent \textbf{Modality Missing in Multi-modal Learning}.
As a widely existing challenge in the realistic scenario, handling modality missing has drawn the attention of the multi-modal learning community. Some early works \cite{pandey2017variational,wu2018multimodal,suzuki2016joint} build their methods based on conditional VAE to capture the multi-modal distribution for the cross-modal generation. \cite{tsai2019multimodal} as one of the recent works utilizes cross-modal fusion to improve the model robustness for modality missing in testing. \cite{shi2020relating} proposes a contrastive framework for learning both paired and unpaired data. In \cite{ma2021smil}, the authors leverage Bayesian meta-learning to reconstruct pseudo text input from image input to resolve the missing modality issue. Instead of training a generative model from scratch, we utilize the large-scaled pre-trained model \cite{li2023blip,li2022blip} and prompt augmentation to achieve effective cross-modal generation for completing the missing data pairs.

\noindent \textbf{Vision-language Pre-training}.
Represented by CLIP \cite{radford2021learning} and ALIGN \cite{jia2021scaling}, the large-scale Vision and Language Pre-training (VLP) models have demonstrated their surprising performance in many downstream vision-language learning tasks\cite{he2019fine} and strong adaptability to new scenarios. A few works have taken the first steps towards incorporating federated learning with pre-training techniques. In \cite{10.1145/3510033}, the authors propose a splitting learning-based framework for training large-scale models like BERT in federated learning systems. PromptFL \cite{guo2022promptfl} allows the clients to train shared soft prompts collaboratively using CLIP \cite{radford2021learning} to provide strong adaptation capability to distributed users tasks. \cite{https://doi.org/10.48550/arxiv.2211.08025,lu2023fedclip,wu2023leveraging,Wu_Huang_Hu_Huang_2023} are trying to explore the efficient methods for lightweight and fast adaptation of pre-trained models. \cite{tan2022federated} proposes FedPCL to transfer shared knowledge among the clients based on prototype contrastive learning. In this work, instead of fine-tuning the large-scale pre-trained models or splitting the model into multiple modules, we conduct effective knowledge transferring to enhance the representation learning performance of a lightweight local module. 

\noindent \textbf{Multi-modal Contrastive Learning}.
Contrastive learning is widely used in the self-supervised learning field, where the learned representations will be assigned to positive and negative samples based on the class belongings.  As for its application in multi-modal learning \cite{li2020unimo,liang2022mind,zolfaghari2021crossclr}, instead of using spatial or temporal transforming to a single instance, the positive pairs are defined as the samples with the same ID or time window. In \cite{zolfaghari2021crossclr}, the authors propose CrossCLR to improve the quality of learned joint embedding from multi-modal data with a novel contrastive loss, which utilizes both inter-modality and intra-modality alignment.  \cite{poklukar2022geometric} extends the multi-modal contrastive learning to efficiently align the cross-modal representations.  Inspired by the predecessors, we adopt a multi-modal contrastive loss to improve the quality of the learned multi-modal joint representations based on the modality-specific representation encoded by the frozen pre-trained models.

\section{Methodology}

\begin{figure*}[t]
    \centering 
    \includegraphics[width=\textwidth]{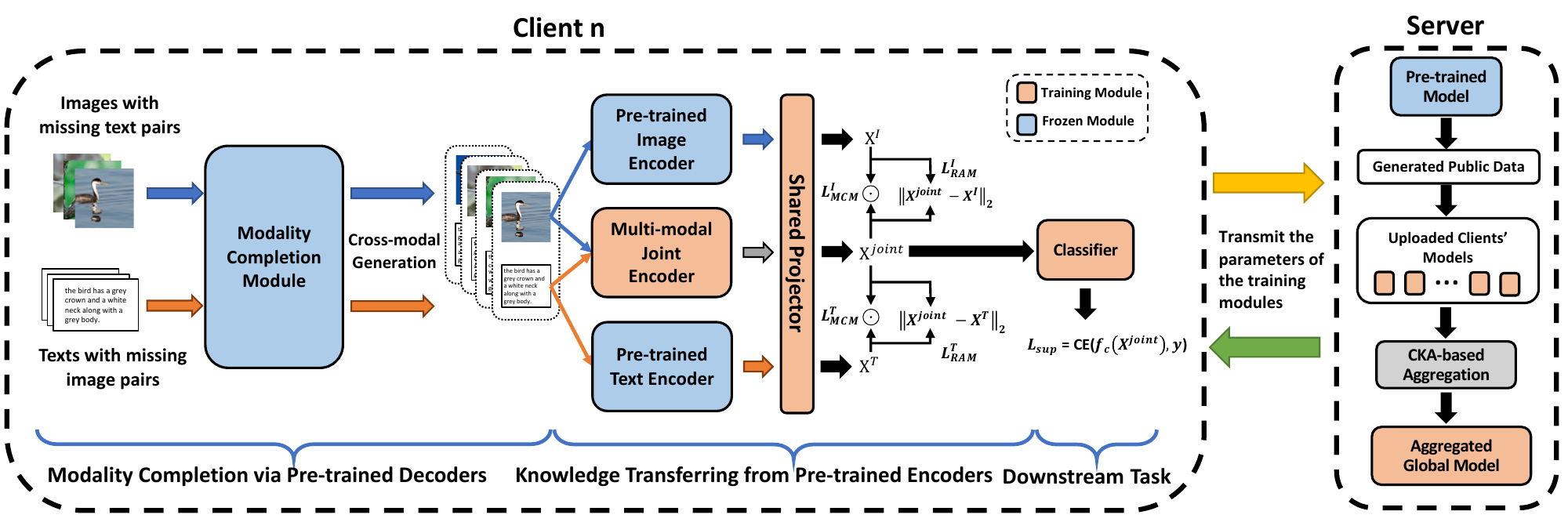}
    \caption{The overview of the proposed FedMVP framework.} 
    \label{Fig.framework} 
\end{figure*}

To explore multi-modal data in federated systems, we propose \textbf{FedMVP} for MFL with the robustness of modality missing during training. As illustrated in Figure~\ref{Fig.framework}, the proposed FedMVP contains four main modules for effective MFL, including \textit{Modality Completion Module}, \textit{Multi-modal Joint Learning Module}, \textit{Knowledge Transferring via Contrastive Training}, \textit{CKA-based Aggregation}.

\subsection{Problem Formulation}
\textbf{Multi-modal Federated Learning.}
In an MFL system, there exist $N$ clients aiming to collaboratively train a global model $w_G$ for multi-modal representation learning. For client $n$, its local data set $\mathcal{D}_n = \{ (X_i,y_i)\}^{|\mathcal{D}_n|}_{i=1}$ contains $|\mathcal{D}_n|$ image-text pairs denoted as $X_i = \{x^I_i, x^T_i\}$, i.e., the $i$-th image data $x^I_i$ and text data $x^T_i$. $y_i$ is the corresponding label. A data instance is denoted as $X_i = \{x^I_i\}$ or $X_i = \{x^T_i\}$ if modality missing happens. Each local model $w_n$ performs on the local task $F_n(\cdot;w_n) : \mathbb{R}^{n} \rightarrow \mathbb{R}^{d} $ and collaborates with other clients for the global task $F_G(\cdot;w_G) : \mathbb{R}^{d_G} \rightarrow \mathbb{R}^{d} $.  Formally, the global objective of MFL for the image-text classification problem is defined as
\begin{equation}
    \min~L_{G}(F_G(\cdot;w_G)) = \min~\sum^{N}_{n=1} \gamma_n L_n(F_n(\mathcal{D}_n; w_n))
\end{equation}
\noindent where $\gamma_n$ is the aggregation weights, and $L_n$ is the local loss function.

\subsection{Local Data Preprocessing}
A pre-trained foundation model is deployed on both the server side and client side, which consists of an image encoder $f^I_{E}(\cdot)$ and a text encoder $f^T_{E}(\cdot)$ for representation extraction, an image decoder $f^I_{D}(\cdot)$ and a text decoder $f^T_{D}(\cdot)$ for the cross-modal generation. 
Notably, all the parameters of the pre-trained models are frozen and will not be transmitted between the server and clients. We will explain the pre-trained model we used below, as well as the details of the local training process.

\noindent\textbf{Modality Completion Module.}
To solve the performance drop problem caused by modality missing, the modality completion module utilizes the cross-modal generation ability of the pre-trained model to complete the missing part of multi-modal data. We use DALLE2 \cite{ramesh2022hierarchical} for text-to-image generation, and BLIP2 \cite{li2023blip} for image-to-text generation. Inspired by \cite{radford2021learning}, we use designed prompts to improve the generation quality of the modality completion module.

\begin{figure*}[tt]
\centering
\subfigure[original text]{
\begin{minipage}[t]{0.2\linewidth}
\centering
\includegraphics[width=\linewidth]{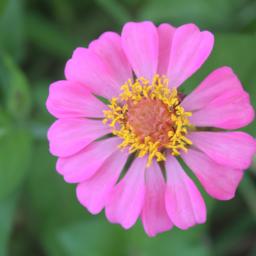} 
\end{minipage}}~~~
% \hspace{1em}
\subfigure[label only]{
\begin{minipage}[t]{0.2\linewidth}
\centering
\includegraphics[width=\linewidth]{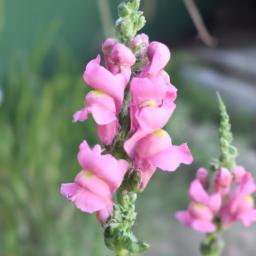}
\end{minipage}}
\subfigure[prompt]{
\begin{minipage}[t]{0.2\linewidth}
\centering
\includegraphics[width=\linewidth]{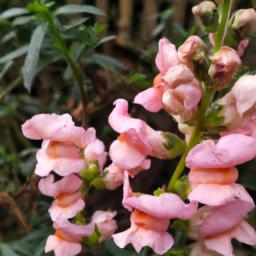}
\end{minipage}}
\subfigure[original image]{
\begin{minipage}[t]{0.2\linewidth}
\centering
\includegraphics[width=\linewidth]{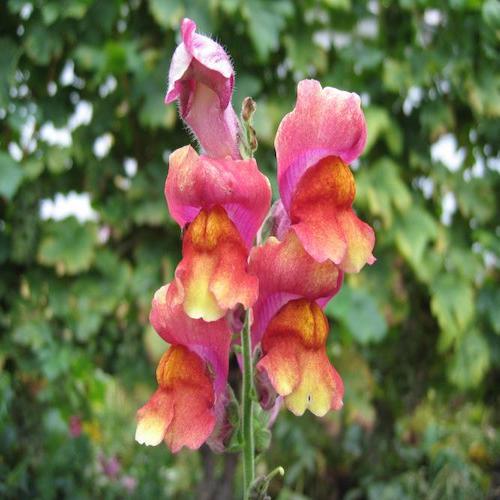}
\end{minipage}}
\centering
% \vskip -1em
\caption{Examples of the generated ``snapdragon'' images.}
\label{fig:prompt_flower}
% \vskip -1em
\end{figure*}

\begin{figure*}[t]
\centering
\subfigure[original text]{
\begin{minipage}[t]{0.2\linewidth}
\centering
\includegraphics[width=\linewidth]{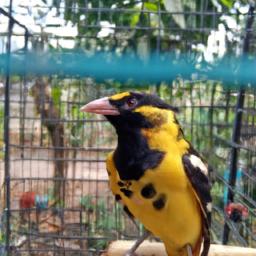} 
\end{minipage}}~~~
% \hspace{1em}
\subfigure[label only]{
\begin{minipage}[t]{0.2\linewidth}
\centering
\includegraphics[width=\linewidth]{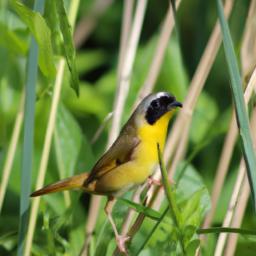}
\end{minipage}}
\subfigure[prompt]{
\begin{minipage}[t]{0.2\linewidth}
\centering
\includegraphics[width=\linewidth]{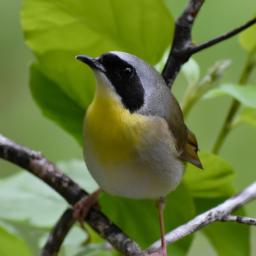}
\end{minipage}}
\subfigure[original image]{
\begin{minipage}[t]{0.2\linewidth}
\centering
\includegraphics[width=\linewidth]{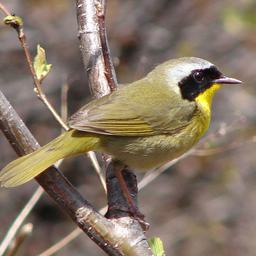}
\end{minipage}}
\centering
% \vskip -1em
\caption{Examples of the generated ``yellowthroat'' images.}
\label{fig:prompt_bird}
% \vskip -1em
\end{figure*}

{\textit{Prompt Augmented Text-to-image Generation.}}
Given an image-text pair $X_i$ with only text data $x^T_i$, the modality completion module could generate an image from a text prompt. To avoid the semantic ambiguities caused by synonyms and polysemy in the text data and label name. Instead of directly using text data $x^T_i$ as the input, we adopt a coarse-to-fine prompt to augment the generation. The prompt template is \textit{``A photo of \{fine-grained label\}, a kind of \{class label\}, \{text description\}''}, which helps the pre-trained models to better understand the characteristics of the generation target and improve the semantic correlation between the text prompt and generated image. 
% We illustrate the generation results with different text inputs in 
% Figure~\ref{fig:prompt_flower} and 
Figures~\ref{fig:prompt_flower} and~\ref{fig:prompt_bird} show examples with different inputs to generate the classes ``snapdragon'' and ``yellow throat'' on the Oxford Flower and CUB-200 datasets, where our designed prompt gives high-quality fake images that are close to the original ones. 

Accordingly, we obtain the image generation prompt based on the original text data, and the process is denoted as $p^T(x^T_i)$. The augmented prompt $p^T(x^T_i)$ will firstly be decomposed by text encoder $f^T_{E}(\cdot)$, then passed to image decoder $f^I_{D}$ for generating the synthetic image $\hat{x}^I_i$, i.e.$\hat{x}^I_i = f^I_{D}(f^T_{E}(p^T(x^T_i)))$.

{\textit{Prompt Augmented Image-to-text Generation.}}
For the image-to-text generation, considering the original text data contains detailed descriptions of the image pair, the direct image captioning result may not be able to cover the fine-grained text details. Therefore, we adapt both the visual question answering (VQA) and image captioning functions of the pre-trained model to generate text pairs $\hat{x}^T_i$ for the image input.  Specifically, with a given image input $x^I_i$, the modality completion module first performs the VQA task over three serial question prompts to get fine-grained descriptions of the image. For instance, given prompt input \textit{``What is the color of the petals?''} for a flower image with the pink pedal, the response answer could be \textit{``Pink''}. After obtaining the answers to the three question prompts, we combine them with the image captioning outcome as the final synthetic text, e.g., \textit{``A photo of \{flower\}, with \{pink\} petals and \{white\} pistils,\{there is a pink flower with a yellow center in the middle of the picture\}''}. We show examples of image-to-text generation in Table~\ref{tab:image_to_text}.

\begin{table*}[t]
\centering
\caption{Image-to-text completion examples from CUB-200 and Oxford Flower.}\label{tab:i2t}
\renewcommand\arraystretch{1.5}
\begin{tabular}{c p{5cm} p{4cm}}
\toprule
\textbf{Type}           & \multicolumn{1}{c}{\textbf{Text}} & \multicolumn{1}{r}{\textbf{Original Image }}         \\ \midrule

Original Text  & this flower is yellow in color, and has petals that are layered.                 & \multirow{2}{*}
{

\begin{minipage}[b]{0.5\columnwidth}
		\centering
		\raisebox{-.5\height}{\includegraphics[width=0.38\linewidth]{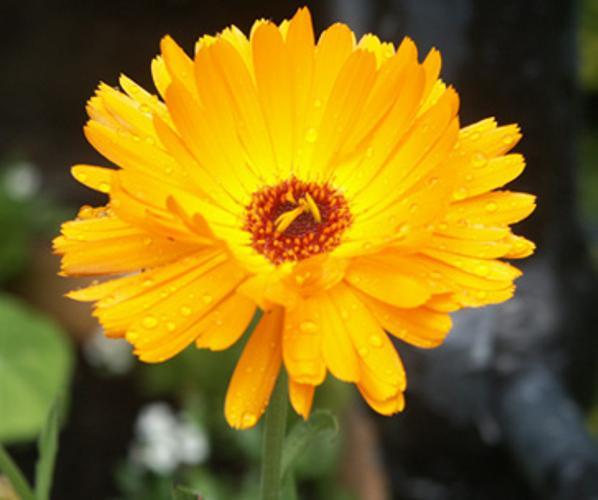}}
	\end{minipage}
 } \\

Synthetic Text & A flower with yellow petals and yellow pistil. yellow flower with water droplets on it in a garden                 &                         \\ \midrule
Original Text  & this bird is yellow and black in color, and has a stubby black beak.                 & \multirow{2}{*}{    \begin{minipage}[b]{0.5\columnwidth}
		\centering
		\raisebox{-.5\height}{\includegraphics[width=0.38\linewidth]{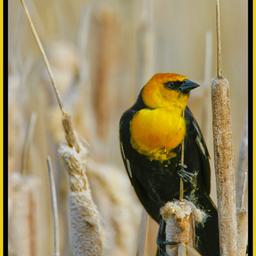}}
	\end{minipage}} \\

Synthetic Text & A bird with black wings and yellow belly. yellow and black bird perched on a cattails plant in a marsh                 &                         \\ \bottomrule
\end{tabular}
\label{tab:image_to_text}
\end{table*}

To better understand the model design and avoid notation confusion, we use completed image-text pair $X_i = \{x^I_i, x^T_i\}$ in the following sections to illustrate how data is processed in FedMVP.

{\textit{Modality-specific Representations.}} 
% The modality-specific representations are produced by the integrated pre-trained model. 
The foundation models are believed to have extraordinary representation extraction ability since they are trained with millions of data instances. Thus, we obtain the image-specific embedding and text-specific embedding via the pre-trained encoders. 
Specifically, we use the {pre-trained Vision Transformer(ViT)} \cite{dosovitskiy2020image} with the patch size of $16 \times 16$ as the image-specific encoder to generate high-quality embedding from image input. The image-specific embedding $\mathbf{X}^I$ is encoded via the pre-trained image encoder $f^I_{E}(\cdot)$ and then mapping to the multi-modal latent space via a shared projection head $f_{shared}(\cdot)$, i.e., $\mathbf{X}^I = f_{shared}(f^I_{E}(\mathbf{x}^I)) \in \mathbb{R}^{d_{latent}}$. Similarly, we get the text-specific embedding $\mathbf{X}^T$ from the {pre-trained BERT model} \cite{devlin2018bert}, where $\mathbf{X}^T = f_{shared}(f^T_{E}(\mathbf{x}^T)) \in \mathbb{R}^{d_{latent}}$. 

\subsection{Local Training}

\noindent\textbf{Multi-modal Joint Learning Module.}
The multi-modal joint learning module contains a joint encoder $f^{Joint}_{E}(\cdot)$ designed to efficiently fuse the image-text information into a complete view. It consists of a cross-modal fusion layer and follows attention-based embedding layers. 

\textit{Pre-processing.} Given an image-text pair $\{{\mathbf{x}^I, \mathbf{x}^T}\}$ as input, we use a non-overlapped patch embedding layer and the pre-trained text encoder $f^T_{E}(\cdot)$ to get the patch sequence $\mathbf{I}_{com}$ and text embedding $\mathbf{T}_{com}$, both belongs to the common dimension $d_{com}$.

\textit{Cross-modal Fusion.} After the positional embedding operation, both the image and text embeddings are fed into the cross-modal fusion layer, which contains a vision-to-language attention module and a language-to-vision attention module. Both modules are based on the cross-modal attention \cite{tsai2019multimodal}, which can effectively fuse the representation between the two input modality embeddings. We take the image-to-text embedding $\mathbf{X}^{I \rightarrow T}$ to show the cross-modal attention:
\begin{equation}
    \begin{split}
        \mathbf{X}^{I \rightarrow T} = CM_{I \rightarrow T}(\mathbf{I}_{com}, \mathbf{T}_{com}) = softmax(\frac{W_{Q_I}\textbf{I}_{com}W^\mathsf{T}_{K_T} \mathbf{T}^\mathsf{T}_{com}}{\sqrt{d_{com}}}) W_{V_T}.
    \end{split}
\end{equation}

Similarly, we can get text-to-image embedding $\mathbf{X}^{T \rightarrow I}$.
The obtained $\mathbf{X}^{I \rightarrow T}$ and $\mathbf{X}^{T \rightarrow I}$ will be concatenated together and projected to the latent space as the final joint embedding via the shared projection head $f_{shared}(\cdot)$ and a self-attention layer as follows:
\begin{equation}
    \mathbf{X}^{joint} =f_{shared}( SelfAttention( \mathbf{X}^{I \rightarrow T} \oplus \mathbf{X}^{T \rightarrow I} )).
\end{equation}

We now obtain the image-specific embedding $\mathbf{X}^I$, text-specific embedding $\mathbf{X}^T$, and joint embedding $\mathbf{X}^{joint}$ in the same latent space $\mathbb{R}^{d_{latent}}$.

\noindent\textbf{Knowledge Transferring from Pre-trained Model.}The training data of large-scale models in the pre-training stage is neither available nor affordable for distributed silos to process, making the fine-tuning and traditional knowledge distillation \cite{hinton2015distilling} of large-scale models impractical under the MFL scenario. In order to transfer the rich representation knowledge from the pre-trained model, we propose \textit{Multi-modal Contrastive Matching (MCM) Loss} and \textit{Representation Aligned Marginal (RAM) Loss} to improve the representation learning performance of the joint encoding module. 

\textit{Multi-modal Contrastive Matching Loss.} To obtain a high-quality joint representation, we utilize the idea of contrastive learning to closer the joint embedding with its corresponding modality-specific embedding and distance it from the embedding of the other categories in the latent space. Let $s_c(x_i,x_j)$ represent the cosine similarity between two embedding, $x_i$ and $x_j$, and $\tau \in (0,1]$ be the temperature hyperparameter. The corresponding scaled similarity is defined as:
$
    sim(x_i,x_j) = \exp(\frac{s_c(x_i, x_j)}{\tau}).
$

Given a batch of embedding $\mathcal{B} = \{\mathbf{X}^T_i, \mathbf{X}^I_i, \mathbf{X}^{joint}_i\}^{|\mathcal{B}|}_{i=1}$, the positive pair for the contrastive learning is defined as the joint embedding with its corresponding modality-specific embedding, i.e., $(\mathbf{X}^T_i, \mathbf{X}^{joint}_i)$ and $(\mathbf{X}^I_i, \mathbf{X}^{joint}_i)$
% \footnote{Recall that only $\mathbf{X}^{joint}_i$ is related to the trainable module, therefore we do not consider $(\mathbf{X}^T_i, \mathbf{X}^{I}_i)$ as part of the positive pair.}
. The other ways of pairing will be treated as negative pairs, denoted as:
\begin{equation}
\begin{split} 
    \Omega^m_i &= \sum\limits_{i \neq j} (sim(\mathbf{X}^M_i, \mathbf{X}^M_j) + sim(\mathbf{X}^M_i, \mathbf{X}^{joint}_j) + sim(\mathbf{X}^{joint}_i, \mathbf{X}^{joint}_j) ),
\end{split}
\end{equation}
\noindent where $M \in \{I, T\}$ indicates the modality type. We define the multi-modal contrastive matching (MCM) loss of all data embedding as follows:
\begin{equation}
\begin{split}
    L_{MCM}(\mathcal{B}) = -\frac{1}{|\mathcal{B}|}\sum^{|\mathcal{B}|}_{i=1} \log\left(\frac{sim(\mathbf{X}^T_i, \mathbf{X}^{joint}_i)}{\Omega^T_i} + \frac{sim(\mathbf{X}^I_i, \mathbf{X}^{joint}_i)}{\Omega^I_i}\right).
\end{split}
\end{equation}

\textit{Representation Aligned Margin Loss.} We propose the \textit{Representation Aligned Margin} (RAM) loss to further enrich the joint representation via pre-trained knowledge to close the semantic gap between the joint embedding and the modality-specific embeddings. 
We use the classification loss derived from the embeddings to evaluate its representation quality. For the $i$-th data sample, the supervised classification loss of one of its corresponding embeddings is denoted as $ L^M_{sup}(i) = CE(f_c(\mathbf{X}^M_{i}), y_i)$. 

Intuitively, embeddings with lower cross-entropy losses contain more informative features from the raw data. With an embedding batch $\mathcal{B}$, the RAM loss aligns joint embedding with image and text embedding separately, if the modality-specific embedding has better representation. Thus, the RAM loss is defined as:
\begin{equation}
    {L}_{RAM}(\mathcal{B}) = \frac{1}{|\mathcal{B}|}\sum^{|\mathcal{B}|}_{i=1} \left({L}^I_{RAM}(i) + {L}^T_{RAM}(i)\right),
\end{equation}
\begin{equation}
{L}^M_{RAM}(I) = \begin{cases}
\lVert \mathbf{X}^{joint}_i - \mathbf{X}^M_i \rVert_2, & \text{if } L^M_{sup}(i) < L^{joint}_{sup}(i) \\
0, & \text{otherwise}
\end{cases},
\end{equation}
where $\mathbf{X}^M_i$ and $\mathbf{X}^{joint}_i$ are all derived from the $i$-th sample in the batch, and $|\mathcal{B}|$ is the batch size. The L2 norm is denoted by $\lVert \cdot \rVert_2$.

\textit{Classification Loss.} 
A two-layer linear classifier $f_C(\cdot)$ will serve as the classifier using only joint embedding as input. The supervised classification loss $L_{sup}$ of client $n$ can be obtained:
\begin{equation}\label{eq:fedavg}
L_{sup}(\mathcal{B}) =  \frac{1}{|\mathcal{B}|} \sum_{i=1}^{|\mathcal{B}|} CE \left(f_C\left(\mathbf{X}^{joint}_{i}; \boldsymbol{\omega}_n\right), y_{i}\right),
\end{equation}
where $f_C(\cdot)$ denotes the classifier model of client $n$, $CE(\cdot)$ is the cross-entropy loss function, and $y_{i}$ is the corresponding label of $i$-th joint embedding $\mathbf{X}^{joint}_{i}$.

\noindent\textbf{\textit{Total Loss.}} The final local training loss of client $k$ in FedMVP is:
\begin{equation}
    L_{local}(\mathcal{D}_k) = L_{sup}(\mathcal{D}_k) +  L_{MCM}(\mathcal{D}_k) + L_{RAM}(\mathcal{D}_k),
\end{equation}
\noindent At each communication round, each client will upload the parameters of the multi-modal joint learning module and classifier to the server for further global aggregation.

\subsection{Server Aggregation}

\begin{figure*}[t]
    \centering 
    \includegraphics[width=\textwidth]{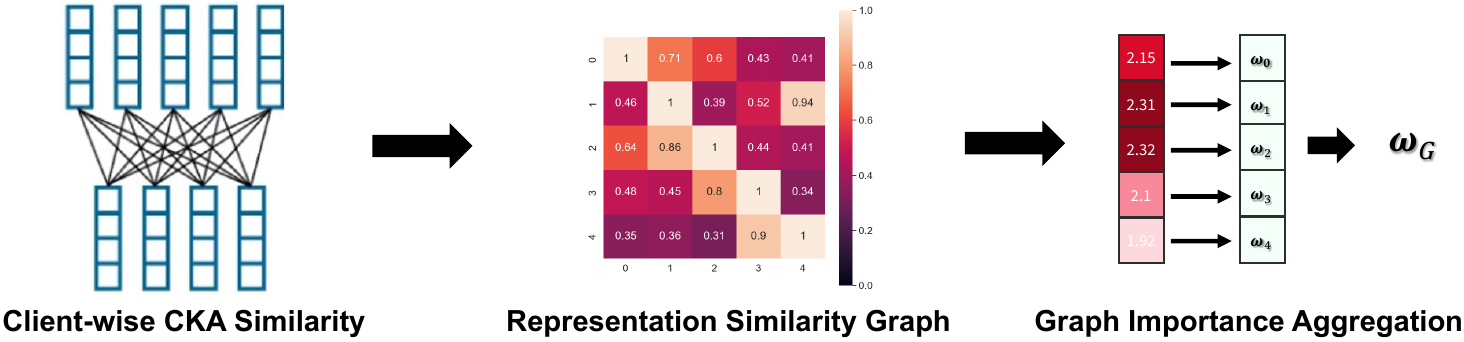}
    \caption{CKA-based Server Aggregation} 
    \label{Fig.aggregation} 
\end{figure*}

Previous works tend to aggregate based on the modality type held by the clients \cite{10.1145/3534678.3539384,yang2022cross}, share public dataset \cite{yu2023multimodal}, or model structure \cite{zhao2022multimodal}, which may lead to data privacy leakage and lacking uniformity. To better enhance the representational ability of the global model, we propose a server aggregation method based on the similarity of model output representations.

At the beginning of the aggregation phase, the server-side pre-trained model will automatically generate $m$ synthetic data pairs $X_m$, where the data amount $m$ is equal to the number of classes of the dataset. Given an uploaded client model, its output representations with generated data are defined as:
\begin{equation}
    \mathbf{X}_{\omega} = [F_{\omega}(X_1), \ldots, F_{\omega}(X_m)]^{T} \in \mathbb{R}^{m \times d_{out}}.
\end{equation}

To  measure the similarity of the model representations among the clients, we utilize the centered kernel alignment (CKA) metric \cite{kornblith2019similarity} based on the output representations from upload models, which is defined as follows:
\begin{equation}
    s_{ij}(\omega_i, \omega_{j}) = \frac{Cov(\mathbf{X}_{\omega_i}, \mathbf{X}_{\omega_j})}{\sqrt{Cov(\mathbf{X}_{\omega_i}, \mathbf{X}_{\omega_i}) 
 Cov(\mathbf{X}_{\omega_j}, \mathbf{X}_{\omega_j})}},
\end{equation}

where $Cov(X,Y) = (m-1)^2 tr(XX^{T}H_{m}YY^{T}H_{m})$, $H_m$ is the centering matrix, $tr(\cdot)$ denotes the matrix trace, $m$ represents the number of input represents. 

With the calculated representation similarity scores, the server constructs a representation similarity graph to illustrate the relationship among clients, as shown in Figure~\ref{Fig.aggregation}. The importance of each client in the representation similarity graph is determined by the sum of its similarity score with all the other clients.
\begin{equation}
    \gamma^t_i = softmax([s_1, \ldots,s_i,\ldots, s_K]),
\end{equation}
where $K$ is the number of clients who participate in the $t$-th aggregation, $s_i = \sum^{K-1}_{j=1}s_{ij}$ is the collection of the graph importance of all $K$ clients. Finally, the global model is weighted and aggregated based on the clients’ graph importance $\gamma^t_i$ as follows: 
\begin{equation}
    w^t_G = \sum^K_{i=1}\gamma^t_i w^t_i.
\end{equation}

\section{Experiments}
% In this section, we demonstrate the performance of the proposed FedMVP on two image-text classification datasets and compare it with both unimodal and multi-modal baselines. 

\subsection{Experiment Setting}

\textbf{Datasets}.
We evaluate the proposed FedMVP on two multi-modal fine-grained categorization datasets, \emph{The Caltech-UCSD Birds-200-2011 (CUB-200) dataset} \cite{WahCUB_200_2011} and \emph{Oxford Flower}\cite{Nilsback08}. Both contain paired image-text data, and each image has $10$ related descriptive text. CUB-200 has $200$ bird classes with $10610$ training image-text instances and $1178$ for testing. Oxford Flower has $102$ flower classes, a training size of $7370$, and a testing size of $819$.

\textbf{Data Distribution Setting}.
For \textbf{Independent Identically Distribution (IID)} setting, we equally distribute the training data to $10$ clients with random selection. Each client will hold the same quantity of local data with a balanced category distribution. To simulate the \textbf{non-IID} scenario in federated systems, we divide the training data set into $C$ shards according to the data set categories, i.e., $200$ shards for CUB-200-2011 dataset and $102$ shards for the Oxford Flower dataset. With fixed $10$ clients, the data shards are randomly and equally distributed to clients. 

% For CUB-200-2011 dataset, each client will receive $20$ shards of data. For Oxford Flower dataset, $8$ clients will hold $10$ data shards and two clients will have $11$ of them.

\textbf{Modality Missing Setting}.
We set $\beta \in [0,1]$ as the missing ratio. For example, given a constant $\beta = 0.3$, $30\%$ randomly selected image-text pairs will lose either image or text data in equal chances. We select $\beta = 0.3, 0.5, 0.8$ to conduct our experiments, and the number of missing images and texts is the same. 

\textbf{Training Setting}.
With fixed $10$ clients, the total communication round is $200$. In each communication round, the clients will perform $10$ epochs for local training with their own local datasets and the server will randomly select $70\%$ of clients for aggregation. We choose AdamW as the optimization function with a scheduler-controlled learning rate $2e-5$. We adopt the warm-up scheduler and cosine annealing scheduler for the training process as well.

\textbf{Baselines}.
Since the existing approaches for addressing modality missing in multi-modal federated learning are relatively limited, we choose \textbf{FedViT}, \textbf{FedBERT} as the uni-modal baseline and \textbf{FedCLIP}, \textbf{FedViLT}, \textbf{MMFed} as the multi-modal baseline. FedViT~\cite{dosovitskiy2020image}, FedBERT~\cite{devlin2018bert}, FedCLIP~\cite{radford2021learning} and FedViLT~\cite{ma2022multimodal} are using large-scale foundation models pre-trained with millions of data as the local models. These large models are fine-tuned on the local data and upload all the parameters to the server for aggregation. MMFed~\cite{xiong2022unified} is a federated multi-modal learning method without leveraging foundation models. FedViLT~\cite{ma2022multimodal} is designed specifically for modality missing. 
\textit{Please refer to Appendix for details of the implementation.}

\begin{table*}[t]
\centering
\caption{\label{tab:iid} Evaluating the impact of incomplete modality on CUB-200 and Oxford Flower datasets under IID setting. $\beta$ indicates the missing ratio of the training set.}
\begin{tabular}{c ccc ccc}

% \caption{IID Experiment on }
\toprule
\multirow{2}{*}{\textbf{Methods}} & \multicolumn{3}{c}{\textbf{CUB-200}}    & \multicolumn{3}{c}{\textbf{Oxford Flower}} \\ \cmidrule(r){2-7} 
                        & $\beta=0.3$ & $\beta=0.5$ & $\beta=0.8$ &  $\beta=0.3$  & $\beta=0.5$ & $\beta=0.8$\\ \midrule
FedViT                       &     $74.71\%$      &   $67.12\%$   &     $60.33\%$      &   $92.15\%$    &    $84.52\%$    &     $76.64\%$      \\
FedBERT                    &   $66.76\%$   &   $58.98\%$     &      $52.54\%$      &  $74.23\%$    &    $70.72\%$     &     $67.81\%$      \\
FedCLIP                   &      $75.73\%$     &  $69.68\%$   &   $63.41\%$       & $91.12\%$   &  $86.32\%$        &    $78.55\%$         \\
FedViLT                      &  $76.29\%$     &   $70.28\%$    &    $64.11\%$        &      $92.67\%$      &   $88.31\%$  &     $81.52\%$    \\ 
MMFed                      &  $63.15\%$     &   $57.48\%$    &    $51.60\%$        &      $72.91\%$      &   $69.43\%$  &     $64.05\%$    \\ \midrule
FedMVP(Ours)            &    $\textbf{77.89\%}$   &   $\textbf{74.46\%}$    &     $\textbf{70.31\%}$      &     $\textbf{93.19\%}$       &   $\textbf{91.28\%}$   &    $\textbf{89.32\%}$   \\ \bottomrule
\end{tabular}

\end{table*}

\begin{table*}[t]
\centering
\caption{Evaluating the impact of incomplete modality on CUB-200 and Oxford Flower datasets under the non-IID setting. $\beta$ indicates the missing ratio of the training set.}
\begin{tabular}{c ccc ccc}
% \caption{IID Experiment on }
\toprule
\multirow{2}{*}{\textbf{Methods}} & \multicolumn{3}{c}{\textbf{CUB-200}}    & \multicolumn{3}{c}{\textbf{Oxford Flower}} \\ \cmidrule(r){2-7} 
                        & $\beta=0.3$ & $\beta=0.5$ & $\beta=0.8$ &  $\beta=0.3$  & $\beta=0.5$ & $\beta=0.8$\\ \midrule
FedViT                       &     $67.05\%$      &   $61.17\%$   &     $50.39\%$      &   $86.25\%$    &    $78.30\%$    &     $70.03\%$      \\
FedBERT                    &   $59.31\%$   &   $51.14\%$     &      $43.67\%$      &  $68.43\%$    &    $62.01\%$     &     $57.16\%$      \\
FedCLIP                   &      $67.63\%$     &  $61.72\%$   &   $56.78\%$       & $85.01\%$   &  $80.13\%$        &    $72.91\%$         \\
FedViLT                      &  $69.19\%$     &   $65.26\%$    &    $58.34\%$        &      $86.96\%$      &   $81.63\%$  &     $73.32\%$    \\ 
MMFed                      &  $57.55\%$     &   $51.12\%$    &    $42.14\%$        &      $65.90\%$      &   $59.26\%$  &     $52.79\%$    \\ \midrule
FedMVP(Ours)            &    $\textbf{72.62\%}$   &   $\textbf{69.73\%}$    &     $\textbf{66.44\%}$      &     $\textbf{88.54\%}$    &   $\textbf{84.78\%}$   &    $\textbf{82.47\%}$   \\ \bottomrule
\end{tabular}
\label{tab:noniid}
\end{table*}

\begin{table*}[t]
\centering
\caption{Evaluating the robustness of the methods over different test sets. \textit{image only} and \textit{text only} indicate the test set only contains either image or text. All the methods are trained over train set WITHOUT modality missing.}\label{tab:test_missing}
\begin{tabular}{c ccc ccc}

% \caption{IID Experiment on }
\toprule
\multirow{2}{*}{\textbf{Methods}} & \multicolumn{3}{c}{\textbf{CUB-200}}    & \multicolumn{3}{c}{\textbf{Oxford Flower}} \\ \cmidrule(r){2-7} 
                        & image only & text only & complete & image only  & text only & complete \\ \midrule
FedCLIP          &  $56.47\%$     &  $47.30\%$    &  $79.73\%$  &  $64.11\%$  &  $53.59\%$  &  $94.12\%$   \\
FedViLT          &  $64.55\%$ &  $52.08\%$  &   $\textbf{82.29\%}$  &  $76.71\%$  &  $60.91\%$  &  $\textbf{96.67\%}$\\ 
MMFed            &  $7.94\%$  &  $13.07\%$  &   $65.28\%$  &  $26.37\%$  &  $40.90\%$ &  $74.89\%$ \\ \midrule
FedMVP(Ours)     &   $\textbf{70.39\%}$  & $\textbf{64.44\%}$ & ${80.79\%}$ & $\textbf{{80.82\%}}$  &     $\textbf{73.50\%}$  &  ${94.27\%}$ \\ \bottomrule
\end{tabular}

\end{table*}

\subsection{Empirical Results}

\textbf{Results of the IID Setting}.
Table \ref{tab:iid} shows the superior performance of FedMVP across different missing ratios under the IID setting on both CUB-200 and Oxford Flower datasets. Observably, all models exhibit a decline in accuracy with an increase in the missing ratio ($\beta$). FedMVP outperforms baseline methods consistently and demonstrates exceptional resilience to performance degradation due to missing modalities. For instance, on the CUB-200 dataset, FedMVP's accuracy margin over the next best-performing model, FedViLT, widens from about $1.6\%$ at $\beta=0.3$ to $6.2\%$ at $\beta=0.8$. A similar trend is observed on the Oxford Flower dataset, with the margin increasing from $0.52\%$ to $7.8\%$. The rate of performance degradation of FedMVP is notably slower than the other models. Specifically, as $\beta$ increases from $0.3$ to $0.8$, the accuracy of FedMVP drops by merely $7.58\%$ and $3.87\%$ on the CUB-200 and Oxford Flower datasets, respectively. In contrast, FedViT witnesses larger drops of $14.38\%$ and $15.51\%$.

\textbf{Results of the Non-IID Setting}.
The non-IID experimental results, presented in Table \ref{tab:noniid}, all methods experience a significant decrease in accuracy compared to the IID setting, including FedMVP. The proposed FedMVP consistently outperforms the other methods across the settings. FedMVP has minimal performance degradation caused by non-IID compared to the baseline methods, with no more than $5\%$ drop on CUB-200 and no more than $7\%$ on Oxford Flower. Despite the increasing missing ratio from $\beta=0.3$ to $\beta=0.8$, FedMVP maintains a substantial lead in accuracy on both datasets. For instance, even with $\beta=0.8$, FedMVP achieves an accuracy of $66.44\%$ and $82.47\%$ on the CUB-200 and Oxford Flower datasets, respectively, confirming its robustness to modality incompleteness under non-IID settings. Notably, the performance margin between FedMVP and baseline is further widened compared to the IID setting. For instance, on the Oxford Flower dataset, as $\beta=0.8$, the accuracy of FedMVP is $29.68\%$ higher than MMFed compared to $25.27\%$ under IID.

\textbf{Results of Single-modality Testing.} Shown in Table~\ref{tab:test_missing}, all methods experience significant performance drops when tested with only one modality (image or text). FedMVP shows the best resilience, achieving the highest accuracy in both image-only and text-only scenarios across datasets. FedViLT\cite{ma2022multimodal} performs best with complete data since it has $26.7\times$ more parameters than FedMVP and is pre-trained over millions of pre-training data. It holds second place in single-modality tests. FedCLIP's performance is limited by local dataset size but benefits from separate ViT and BERT encodings. MMFed suffers the most due to its co-attention mechanism and performs better in text-only testing due to its integrated BERT. In summary, FedMVP demonstrates robustness in both training and testing under missing modalities.

\textbf{Ablation Study.} The results in Table \ref{tab:ablation} show that all the modules in the FedMVP model significantly contribute to its performance. Experimental results show that MCM loss and RAM loss can effectively improve the quality of the representation generated by the multi-modal joint encoder and enhance the final performance of the model by transferring pre-trained knowledge through representation learning. The modality completion module can supplement the data by providing additional training information using the transferable knowledge of the pre-trained model. Furthermore, the experimental results suggest that CKA similarity can effectively measure the importance of the representation learned by each client's local model and can improve aggregation performance compared to traditional average aggregation.

\begin{table}[t]

\centering
\caption{\label{tab:ablation}Ablation study on both CUB-200 and Oxford Flower datasets with $\beta=0.3$ under non-IID setting; wo/MCM denoting MCM Loss excluded; wo/RAM excludes RAM loss; wo/Completion refers to training without modality completion module; wo/CKA indicates server aggregation as FedAvg. }
\begin{tabular}{l c c}
\toprule
\multicolumn{1}{c}{\textbf{Model}}  & \textbf{CUB-200} & \textbf{Oxford Flower} \\ \midrule
\multicolumn{1}{l}{\textbf{FedMVP}} &         $72.62\%$         &              $88.54\%$          \\ \midrule
-wo/MCM    &         $66.87\%$         &      $81.44\%$    \\ 
-wo/RAM &      $68.25\%$     &      $83.60\%$   \\ 
-wo/Completion  & $67.49\%$   &  $81.87\%$    \\ 
-wo/CKA  &   $70.11\%$       &    $85.01\%$   \\ \bottomrule
\end{tabular}
\end{table}

\section{Conclusion}
In conclusion, we proposed the FedMVP framework to tackle modality missing, a widely existing real-world challenge, where part of the multi-modal data is incomplete and unaligned. Our framework utilizes large-scale pre-trained models with frozen parameters for modality completion and representation knowledge transfer at each client. It provides a solution for integrating large-scale pre-trained models to empower the federated system with robustness towards modality incompleteness. The experiments on the real-world image-text pair benchmark demonstrated the effectiveness of our proposed method. The proposed FedMVP framework shows great potential in addressing the missing modality and unified representation learning challenges of multi-modal federated learning. We hope this work can provide inspiration for future research in this field.

\begin{credits}
\subsubsection{\ackname} This study was partially supported by the National Science Foundation under Grant No. 2238275, 2333790, and 2348541.
\end{credits}


\begin{thebibliography}{10}
\providecommand{\url}[1]{\texttt{#1}}
\providecommand{\urlprefix}{URL }
\providecommand{\doi}[1]{https://doi.org/#1}

\bibitem{baltruvsaitis2018multimodal}
Baltru{\v{s}}aitis, T., Ahuja, C., Morency, L.P.: Multimodal machine learning: A survey and taxonomy. IEEE transactions on pattern analysis and machine intelligence  \textbf{41}(2),  423--443 (2018)

\bibitem{9671374}
Che, L., Long, Z., Wang, J., Wang, Y., Xiao, H., Ma, F.: Fedtrinet: A pseudo labeling method with three players for federated semi-supervised learning. In: 2021 IEEE Big Data. pp. 715--724 (2021)

\bibitem{s23156986}
Che, L., Wang, J., Zhou, Y., Ma, F.: Multimodal federated learning: A survey. Sensors  \textbf{23}(15) (2023)

\bibitem{devlin2018bert}
Devlin, J., Chang, M.W., Lee, K., Toutanova, K.: Bert: Pre-training of deep bidirectional transformers for language understanding. arXiv:1810.04805  (2018)

\bibitem{dosovitskiy2020image}
Dosovitskiy, A., Beyer, L., Kolesnikov, A., Weissenborn, D., Zhai, X., Unterthiner, T., Dehghani, M., Minderer, M., Heigold, G., Gelly, S., et~al.: An image is worth 16x16 words: Transformers for image recognition at scale. arXiv preprint arXiv:2010.11929  (2020)

\bibitem{kairouz2021advances}
Kairouz, P., McMahan, H.B., Avent, B., Bellet, A., Bennis, M., Bhagoji, A.N., Bonawitz, K., Charles, Z., Cormode, G., Cummings, R., et~al.: Advances and open problems in federated learning. Foundations and trends{\textregistered} in machine learning  \textbf{14}(1--2),  1--210 (2021)

\bibitem{kim2021vilt}
Kim, W., Son, B., Kim, I.: Vilt: Vision-and-language transformer without convolution or region supervision. In: International Conference on Machine Learning. pp. 5583--5594. PMLR (2021)

\bibitem{10233897}
Lan, G., Liu, X.Y., Zhang, Y., Wang, X.: Communication-efficient federated learning for resource-constrained edge devices. IEEE Transactions on Machine Learning in Communications and Networking  \textbf{1},  210--224 (2023)

\bibitem{NEURIPS2023_bc6a1f96}
Lan, G., Wang, H., Anderson, J., Brinton, C., Aggarwal, V.: Improved communication efficiency in federated natural policy gradient via admm-based gradient updates. In: Oh, A., Naumann, T., Globerson, A., Saenko, K., Hardt, M., Levine, S. (eds.) Advances in Neural Information Processing Systems. vol.~36, pp. 59873--59885. Curran Associates, Inc. (2023)

\bibitem{liang2021omnilytics}
Liang, J., Li, S., Cao, B., Jiang, W., He, C.: Omnilytics: A blockchain-based secure data market for decentralized machine learning (2021)

\bibitem{long2021fedsiam}
Long, Z., Che, L., Wang, Y., Ye, M., Luo, J., Wu, J., Xiao, H., Ma, F.: Fedsiam: Towards adaptive federated semi-supervised learning (2021)

\bibitem{loshchilov2017decoupled}
Loshchilov, I., Hutter, F.: Decoupled weight decay regularization. arXiv preprint arXiv:1711.05101  (2017)

\bibitem{Lyu2023-cd}
Lyu, W., Dong, X., Wong, R., Zheng, S., Abell-Hart, K., Wang, F., Chen, C.: A multimodal transformer: Fusing clinical notes with structured {EHR} data for interpretable {In-Hospital} mortality prediction. {AMIA}. Annual Symposium proceedings. AMIA Symposium  \textbf{2022},  719--728 (2023)

\bibitem{https://doi.org/10.1002/alz.068774}
Ma, H., Liu, Y., Wu, G.: Elucidating multi-stage progression of neuro-degeneration process in alzheimer’s disease. Alzheimer's \& Dementia  \textbf{18}(S6),  e068774 (2022)

\bibitem{MA2024120609}
Ma, H., Shi, Z., Kim, M., Liu, B., Smith, P.J., Liu, Y., Wu, G.: Disentangling sex-dependent effects of apoe on diverse trajectories of cognitive decline in alzheimer's disease. NeuroImage  \textbf{292},  120609 (2024)

\bibitem{ma2022multimodal}
Ma, M., Ren, J., Zhao, L., Testuggine, D., Peng, X.: Are multimodal transformers robust to missing modality? In: Proceedings of the IEEE/CVF Conference on Computer Vision and Pattern Recognition. pp. 18177--18186 (2022)

\bibitem{10198520}
Mawuli, C.B., Che, L., Kumar, J., Din, S.U., Qin, Z., Yang, Q., Shao, J.: Fedstream: Prototype-based federated learning on distributed concept-drifting data streams. IEEE Transactions on Systems, Man, and Cybernetics: Systems  \textbf{53}(11),  7112--7124 (2023)

\bibitem{radford2021learning}
Radford, A., Kim, J.W., Hallacy, C., Ramesh, A., Goh, G., Agarwal, S., Sastry, G., Askell, A., Mishkin, P., Clark, J., et~al.: Learning transferable visual models from natural language supervision. In: International conference on machine learning. pp. 8748--8763. PMLR (2021)

\bibitem{10446126}
Ruan, K., He, X., Wang, J., Zhou, X., Feng, H., Kebarighotbi, A.: S2e: Towards an end-to-end entity resolution solution from acoustic signal. In: ICASSP 2024 - 2024 IEEE International Conference on Acoustics, Speech and Signal Processing (ICASSP). pp. 10441--10445 (2024)

\bibitem{9992745}
Wang, H., Marella, S., Anderson, J.: Fedadmm: A federated primal-dual algorithm allowing partial participation. In: 2022 IEEE 61st Conference on Decision and Control (CDC). pp. 287--294 (2022)

\bibitem{pmlr-v211-wang23d}
Wang, H., Toso, L.F., Anderson, J.: Fedsysid: A federated approach to sample-efficient system identification. In: Matni, N., Morari, M., Pappas, G.J. (eds.) Proceedings of The 5th Annual LDCC. Proceedings of Machine Learning Research, vol.~211, pp. 1308--1320. PMLR (15--16 Jun 2023)

\bibitem{wang2024towards}
Wang, J., Yang, X., Cui, S., Che, L., Lyu, L., Xu, D.D., Ma, F.: Towards personalized federated learning via heterogeneous model reassembly. Advances in Neural Information Processing Systems  \textbf{36} (2024)

\bibitem{wang2023knowledge}
Wang, J., Zeng, S., Long, Z., Wang, Y., Xiao, H., Ma, F.: Knowledge-enhanced semi-supervised federated learning for aggregating heterogeneous lightweight clients in iot. In: Proceedings of the 2023 SIAM International Conference on Data Mining (SDM). pp. 496--504. SIAM (2023)

\bibitem{xiong2022unified}
Xiong, B., Yang, X., Qi, F., Xu, C.: A unified framework for multi-modal federated learning. Neurocomputing  \textbf{480},  110--118 (2022)

\end{thebibliography}


\begin{thebibliography}{10}
\providecommand{\url}[1]{\texttt{#1}}
\providecommand{\urlprefix}{URL }
\providecommand{\doi}[1]{https://doi.org/#1}

\bibitem{9671374}
Che, L., Long, Z., Wang, J., Wang, Y., Xiao, H., Ma, F.: Fedtrinet: A pseudo labeling method with three players for federated semi-supervised learning. In: 2021 IEEE Big Data. pp. 715--724 (2021)

\bibitem{s23156986}
Che, L., Wang, J., Zhou, Y., Ma, F.: Multimodal federated learning: A survey. Sensors  \textbf{23}(15) (2023)

\bibitem{10.1145/3534678.3539384}
Chen, J., Zhang, A.: Fedmsplit: Correlation-adaptive federated multi-task learning across multimodal split networks. In: ACM SIGKDD. p. 87–96 (2022)

\bibitem{https://doi.org/10.48550/arxiv.2211.08025}
Chen, J., Xu, W., Guo, S., Wang, J., Zhang, J., Wang, H.: Fedtune: A deep dive into efficient federated fine-tuning with pre-trained transformers (2022)

\bibitem{chen2022towards}
Chen, S., Li, B.: Towards optimal multi-modal federated learning on non-iid data with hierarchical gradient blending. In: IEEE INFOCOM 2022-IEEE Conference on Computer Communications. pp. 1469--1478. IEEE (2022)

\bibitem{COBBINAH2022102585}
Cobbinah, B.M., Sorg, C., Yang, Q., Ternblom, A., Zheng, C., Han, W., Che, L., Shao, J.: Reducing variations in multi-center alzheimer’s disease classification with convolutional adversarial autoencoder. Medical image analysis  \textbf{82},  102585 (2022)

\bibitem{devlin2018bert}
Devlin, J., Chang, M.W., Lee, K., Toutanova, K.: Bert: Pre-training of deep bidirectional transformers for language understanding. arXiv:1810.04805  (2018)

\bibitem{dosovitskiy2020image}
Dosovitskiy, A., Beyer, L., Kolesnikov, A., Weissenborn, D., Zhai, X., Unterthiner, T., Dehghani, M., Minderer, M., Heigold, G., Gelly, S., et~al.: An image is worth 16x16 words: Transformers for image recognition at scale. arXiv preprint arXiv:2010.11929  (2020)

\bibitem{guo2022promptfl}
Guo, T., Guo, S., Wang, J., Xu, W.: Promptfl: Let federated participants cooperatively learn prompts instead of models--federated learning in age of foundation model. arXiv preprint arXiv:2208.11625  (2022)

\bibitem{he2019fine}
He, X., Peng, Y.: Fine-grained visual-textual representation learning. IEEE Transactions on Circuits and Systems for Video Technology  \textbf{30}(2),  520--531 (2019)

\bibitem{hinton2015distilling}
Hinton, G., Vinyals, O., Dean, J.: Distilling the knowledge in a neural network (2015)

\bibitem{jia2021scaling}
Jia, C., Yang, Y., Xia, Y., Chen, Y.T., Parekh, Z., Pham, H., Le, Q., Sung, Y.H., Li, Z., Duerig, T.: Scaling up visual and vision-language representation learning with noisy text supervision. In: International Conference on Machine Learning. pp. 4904--4916. PMLR (2021)

\bibitem{kornblith2019similarity}
Kornblith, S., Norouzi, M., Lee, H., Hinton, G.: Similarity of neural network representations revisited. In: International Conference on Machine Learning. pp. 3519--3529. PMLR (2019)

\bibitem{li2023blip}
Li, J., Li, D., Savarese, S., Hoi, S.: Blip-2: Bootstrapping language-image pre-training with frozen image encoders and large language models. arXiv preprint arXiv:2301.12597  (2023)

\bibitem{li2022blip}
Li, J., Li, D., Xiong, C., Hoi, S.: Blip: Bootstrapping language-image pre-training for unified vision-language understanding and generation. In: International Conference on Machine Learning. pp. 12888--12900. PMLR (2022)

\bibitem{li2020unimo}
Li, W., Gao, C., Niu, G., Xiao, X., Liu, H., Liu, J., Wu, H., Wang, H.: Unimo: Towards unified-modal understanding and generation via cross-modal contrastive learning. arXiv preprint arXiv:2012.15409  (2020)

\bibitem{liang2022mind}
Liang, W., Zhang, Y., Kwon, Y., Yeung, S., Zou, J.: Mind the gap: Understanding the modality gap in multi-modal contrastive representation learning. arXiv preprint arXiv:2203.02053  (2022)

\bibitem{liu2020federated}
Liu, F., Wu, X., Ge, S., Fan, W., Zou, Y.: Federated learning for vision-and-language grounding problems. In: Proceedings of the AAAI Conference on Artificial Intelligence. vol.~34, pp. 11572--11579 (2020)

\bibitem{lu2023fedclip}
Lu, W., Hu, X., Wang, J., Xie, X.: Fedclip: Fast generalization and personalization for clip in federated learning. arXiv preprint arXiv:2302.13485  (2023)

\bibitem{Ma_2022_CVPR}
Ma, M., Ren, J., Zhao, L., Testuggine, D., Peng, X.: Are multimodal transformers robust to missing modality? In: Proceedings of the IEEE/CVF Conference on Computer Vision and Pattern Recognition (CVPR). pp. 18177--18186 (June 2022)

\bibitem{ma2022multimodal}
Ma, M., Ren, J., Zhao, L., Testuggine, D., Peng, X.: Are multimodal transformers robust to missing modality? In: Proceedings of the IEEE/CVF Conference on Computer Vision and Pattern Recognition. pp. 18177--18186 (2022)

\bibitem{ma2021smil}
Ma, M., Ren, J., Zhao, L., Tulyakov, S., Wu, C., Peng, X.: Smil: Multimodal learning with severely missing modality. In: Proceedings of the AAAI Conference on Artificial Intelligence. vol.~35, pp. 2302--2310 (2021)

\bibitem{mcmahan2017communication}
McMahan, B., Moore, E., Ramage, D., Hampson, S., y~Arcas, B.A.: Communication-efficient learning of deep networks from decentralized data. In: Artificial intelligence and statistics. pp. 1273--1282. PMLR (2017)

\bibitem{Nilsback08}
Nilsback, M.E., Zisserman, A.: Automated flower classification over a large number of classes. In: Indian Conference on Computer Vision, Graphics and Image Processing (Dec 2008)

\bibitem{pandey2017variational}
Pandey, G., Dukkipati, A.: Variational methods for conditional multimodal deep learning. In: 2017 international joint conference on neural networks (IJCNN). pp. 308--315. IEEE (2017)

\bibitem{poklukar2022geometric}
Poklukar, P., Vasco, M., Yin, H., Melo, F.S., Paiva, A., Kragic, D.: Geometric multimodal contrastive representation learning. In: International Conference on Machine Learning. pp. 17782--17800. PMLR (2022)

\bibitem{radford2021learning}
Radford, A., Kim, J.W., Hallacy, C., Ramesh, A., Goh, G., Agarwal, S., Sastry, G., Askell, A., Mishkin, P., Clark, J., et~al.: Learning transferable visual models from natural language supervision. In: International conference on machine learning. pp. 8748--8763. PMLR (2021)

\bibitem{ramesh2022hierarchical}
Ramesh, A., Dhariwal, P., Nichol, A., Chu, C., Chen, M.: Hierarchical text-conditional image generation with clip latents. arXiv:2204.06125  (2022)

\bibitem{shi2020relating}
Shi, Y., Paige, B., Torr, P.H., Siddharth, N.: Relating by contrasting: A data-efficient framework for multimodal generative models. arXiv preprint arXiv:2007.01179  (2020)

\bibitem{suzuki2016joint}
Suzuki, M., Nakayama, K., Matsuo, Y.: Joint multimodal learning with deep generative models. arXiv preprint arXiv:1611.01891  (2016)

\bibitem{tan2022federated}
Tan, Y., Long, G., Ma, J., Liu, L., Zhou, T., Jiang, J.: Federated learning from pre-trained models: A contrastive learning approach. arXiv:2209.10083  (2022)

\bibitem{10.1145/3510033}
Tian, Y., Wan, Y., Lyu, L., Yao, D., Jin, H., Sun, L.: Fedbert: When federated learning meets pre-training. ACM Trans. Intell. Syst. Technol.  \textbf{13}(4) (2022)

\bibitem{tsai2019multimodal}
Tsai, Y.H.H., Bai, S., Liang, P.P., Kolter, J.Z., Morency, L.P., Salakhutdinov, R.: Multimodal transformer for unaligned multimodal language sequences. In: Proceedings of the conference. Association for Computational Linguistics. Meeting. vol.~2019, p.~6558. NIH Public Access (2019)

\bibitem{WahCUB_200_2011}
Wah, C., Branson, S., Welinder, P., Perona, P., Belongie, S.: Caltech-ucsd birds-200-2011 (cub-200-2011). Tech. rep. (2011)

\bibitem{wang2024rethinking}
Wang, J., Chen, Y., Wu, Y., Das, M., Yang, H., Ma, F.: Rethinking personalized federated learning with clustering-based dynamic graph propagation. In: Pacific-Asia Conference on Knowledge Discovery and Data Mining. pp. 155--167 (2024)

\bibitem{wang2022towards}
Wang, J., Qian, C., Cui, S., Glass, L., Ma, F.: Towards federated covid-19 vaccine side effect prediction. In: Joint European Conference on Machine Learning and Knowledge Discovery in Databases. pp. 437--452. Springer (2022)

\bibitem{wang2024towards}
Wang, J., Yang, X., Cui, S., Che, L., Lyu, L., Xu, D.D., Ma, F.: Towards personalized federated learning via heterogeneous model reassembly. Advances in Neural Information Processing Systems  \textbf{36} (2024)

\bibitem{wang2023knowledge}
Wang, J., Zeng, S., Long, Z., Wang, Y., Xiao, H., Ma, F.: Knowledge-enhanced semi-supervised federated learning for aggregating heterogeneous lightweight clients in iot. In: Proceedings of the 2023 SIAM International Conference on Data Mining (SDM). pp. 496--504. SIAM (2023)

\bibitem{wu2018multimodal}
Wu, M., Goodman, N.: Multimodal generative models for scalable weakly-supervised learning. Advances in neural information processing systems  \textbf{31} (2018)

\bibitem{Wu_Huang_Hu_Huang_2023}
Wu, X., Huang, F., Hu, Z., Huang, H.: Faster adaptive federated learning. Proceedings of the AAAI Conference on Artificial Intelligence  \textbf{37}(9),  10379--10387 (2023)

\bibitem{wu2023leveraging}
Wu, X., Lin, W.Y., Willmott, D., Condessa, F., Huang, Y., Li, Z., Ganesh, M.R.: Leveraging foundation models to improve lightweight clients in federated learning (2023)

\bibitem{xiong2022unified}
Xiong, B., Yang, X., Qi, F., Xu, C.: A unified framework for multi-modal federated learning. Neurocomputing  \textbf{480},  110--118 (2022)

\bibitem{yang2022cross}
Yang, X., Xiong, B., Huang, Y., Xu, C.: Cross-modal federated human activity recognition via modality-agnostic and modality-specific representation learning  (2022)

\bibitem{yu2023multimodal}
Yu, Q., Liu, Y., Wang, Y., Xu, K., Liu, J.: Multimodal federated learning via contrastive representation ensemble. In: ICLR (2023)

\bibitem{zhao2022multimodal}
Zhao, Y., Barnaghi, P., Haddadi, H.: Multimodal federated learning on iot data. In: 2022 IEEE/ACM Seventh International Conference on Internet-of-Things Design and Implementation (IoTDI). pp. 43--54. IEEE (2022)

\bibitem{DBLP:conf/cikm/Zhou0WH22}
Zhou, Y., Wu, J., Wang, H., He, J.: Adversarial robustness through bias variance decomposition: {A} new perspective for federated learning. In: CIKM. pp. 2753--2762. {ACM} (2022)

\bibitem{zolfaghari2021crossclr}
Zolfaghari, M., Zhu, Y., Gehler, P., Brox, T.: Crossclr: Cross-modal contrastive learning for multi-modal video representations. In: Proceedings of the IEEE/CVF International Conference on Computer Vision. pp. 1450--1459 (2021)

\bibitem{zong2021fedcmr}
Zong, L., Xie, Q., Zhou, J., Wu, P., Zhang, X., Xu, B.: Fedcmr: Federated cross-modal retrieval. In: Proceedings of the 44th International ACM SIGIR Conference on Research and Development in Information Retrieval. pp. 1672--1676 (2021)

\end{thebibliography}
\end{document}

\title{Leveraging Foundation Models for Multi-modal Federated Learning with Incomplete Modality (\textcolor{red}{Supplementary Material})}

\titlerunning{Multi-modal Federated Learning with Incomplete Modality}

\author{Liwei Che\inst{1,2} \and
Jiaqi Wang\inst{2} \and
Xinyue Liu\inst{3} \and
Fenglong Ma\inst{2} \corr
}
\authorrunning{L. Che et al.}

\institute{Rutgers University, Piscataway NJ 08854, USA\\ \email{lw.che@rutgers.edu}
\and
Pennsylvania State University, University Park PA 16802, USA \email{\{jawang,fenglong\}@psu.edu}
\and
Dalian University of Technology, Dalian Liaoning 116621, China
\email{xyliu@dlut.edu.cn}}

\maketitle              % typeset the header of the contribution

\appendix

\section{Datasets}

We evaluate the effectiveness of FedMVP with two multi-modal classification datasets, the Caltech-UCSD Birds-200-2011 (CUB-200) dataset and Oxford 102 Flower dataset, shown in Table~\ref{tab:dataset}.

\paragraph{\textbf{The Caltech-UCSD Birds-200-2011 (CUB-200) Dataset}} CUB-200 is an image-text multi-modal dataset for fine-grained visual categorization tasks. It contains $11,788$ images of $200$ subcategories belonging to birds. We split the dataset by $10,610$ for training and $1,178$ for testing. Each image has $10$ pieces of detailed text descriptions.

\paragraph{\textbf{Oxford 102 Flower Dataset}} Oxford Flower is an image-text classification dataset comprising 102 flower categories. Each class consists of between $40$ and $258$ images. The dataset is split into a training set with $7,370$ data samples and a test set with $819$ data samples. Each image has $10$ corresponding text descriptions.

\begin{table*}[!h]
  \centering
  \caption{Dataset details.}
  \label{tab:dataset}
  \begin{tabular}{cccccc}
    \toprule
    \textbf{Dataset} & \textbf{Modality Types} & \textbf{Classes} & \textbf{Train Size} & \textbf{Test Size} & \textbf{Image Size} \\
    \midrule
    CUB-200          & Image and Text          & $200$              & $10610$               & $1178$               & $256 \times 256$ \\
    Oxford Flower    & Image and Text          & $102$              & $7370$                & $819$                & $256 \times 256$ \\
    \bottomrule
  \end{tabular}
\end{table*}

\begin{table*}[!t]
\centering
\caption{The template of text-to-image generation prompt on CUB-200 and Flower102.}
\label{tab:t2i prompt}
\begin{tabularx}{\textwidth}{c p{5cm} p{4.5cm}}
\toprule
% \hline
\textbf{Input Prompt}                & \multicolumn{1}{c}{\textbf{CUB-200}}                                    & \multicolumn{1}{c}{\textbf{Oxford Flower}}                                 \\ \midrule
Example 1                           & A photo of \{common yellowthroat\}, a kind of \{bird\}, \{original text\}.                                      & A photo of \{bird of paradise\}, a kind of \{flower\}, \{original text\}.                                         \\ \midrule
Example 2                           & A photo of \{cardinal\}, a kind of \{bird\}, \{original text\}.                             & A photo of \{buttercup\}, a kind of \{bird\}, \{original text\}.                                                                      \\  \bottomrule
\end{tabularx}

\end{table*}

\section{Prompt-Augmented Modality Completion}

In this section, we provide more details of our modality completion module. 

\subsection{Text-to-image Generation}
We use a coarse-to-fine prompt design to help the pre-trained model generate high-quality synthetic images. The coarse-to-fine prompt can avoid the semantic ambiguities caused by synonyms and polysemy in the text data and label names and improve the generation quality. For instance, a class label \textit{``flower''} can avoid the model identifying \textit{``bird of paradise''} as a bird. Detailed, we adopt the prompt template as the examples shown in \ref{tab:t2i prompt} for the input of the DALLE model provided by OpenAI to generate the synthetic images based on the prompts.

\subsection{Image-to-text Generation}

As shown in Table~\ref{tab:i2t prompt}, after obtaining the answers to the three question prompts, we concatenated the detailed answers with the image captioning outcome to generate the final complete image-text pair. 

\begin{table*}[!h]
\centering
\caption{The template of image-to-text generation prompt on CUB-200 and Flower102.}
\label{tab:i2t prompt}
\begin{tabularx}{\textwidth}{c p{4.5cm} p{4.5cm}}
\toprule
% \hline
\textbf{Input Prompt}                & \multicolumn{1}{c}{\textbf{CUB-200}}                                    & \multicolumn{1}{c}{\textbf{Oxford Flower}}                                 \\ \midrule
Question 1                           & What is in the photo?                                         & What is in the photo?                                            \\ \midrule
Question 2                           & What is the color of the wings?                             & What is the color of the petals?                                \\ \midrule
Question 3                           & What is the color of the belly?                       & What is the color of the pistil?                                \\ \midrule
Image Captioning                     & A photo of - Caption                                                     & A photo of - Caption                                                       \\ 
\midrule
% \hline
\multicolumn{1}{c}{\textbf{Output}} & A \{Answer1\} with \{Answer2\} wings and \{Answer3\} belly. \{Caption\}. & A \{Answer1\} with \{Answer2\} petals and \{Answer3\} pistil. \{Caption\}. \\ \bottomrule
\end{tabularx}

\end{table*}

\section{Implementation Details}
\label{sec:implementation}
\subsection{Data Pre-processing}
We provide the detailed image augmentations we used in Table~\ref{tab:IMGaug}.

\begin{table}[!h]
\centering
\caption{\label{tab:IMGaug} Used image  augmentations.}
\begin{tabular}{l l l}
\toprule
\multicolumn{1}{c}{\textbf{Mode}}  & \textbf{Augmentation} & \textbf{Parameters} \\ \midrule
\multirow{7}{*}{Train} 
& Resize & $224 \times 224$ \\
       & RandomRotate & $45^{\circ}$ \\
        & RandomVerticalFlip & $p=0.5$ \\
      & RandomHorizontalFlip & $p=0.5$ \\
      & RandomGrayscale & $p=0.025$ \\
      & RandomColorJitter & $p=0.8$ \\
      
      & Normalize & - \\ \midrule
\multirow{3}{*}{Test}    
& Resize & $224 \times 224$ \\
        & CenterCrop & input size \\
        & Normalize & - \\
\bottomrule
\end{tabular}

\end{table}

\subsection{Implementation of Modality Missing}
In practice, we use dummy sentence \textit{"This is a photo."} as the placeholder for the text input during the image-only testing, along with a corresponding attention mask. For the text-only setting, we use random Gaussian noise to generate dummy images as the placeholder of the image input, with related attention masks helping focus on text input.

\subsection{Model Implementation}

\subsubsection{Implementation of FedViT}
FedViT~\cite{dosovitskiy2020image} equips each client with pre-trained ViT as the local image encoder with an MLP as the classifier. Both the encoder and classifier are trainable.
\paragraph{Architecture}
The local model consists of a pre-trained $vit\_b\_16$ as image encoder provided by \textit{torchvision} and a three-layer MLP for classification with embedding dimensions of ($1,000$, $768$, $512$, $num\_classes$). Here $num\_classes$ indicates the number of classes of the datasets, i.e., $200$ for CUB-200 and $102$ for Oxford Flower.

\paragraph{Training} The FedViT is trained with AdamW optimize \cite{loshchilov2017decoupled} with a learning rate of $2e-5$. The uploaded model will be aggregated by FedAvg.The total communication round is $200$, with local epochs as $10$. The batch size is $64$.

\subsubsection{Implementation of FedBERT}
Similar to FedViT, FedBERT allows each client to use pre-trained BERT\cite{devlin2018bert} as the local encoder. The local training process is fine-tuning the whole pre-trained BERT model over the local datasets.

\paragraph{Architecture} The local model consists of a pre-trained $BERT-base$ model provided by huggingface and a linear layer for classification with embedding dimensions of $(768, num\_classes)$. 

\paragraph{Training}The FedBERT is trained with AdamW\cite{loshchilov2017decoupled} optimizer with a learning rate of $2e-5$. The uploaded model will be aggregated by FedAvg.The total communication round is $200$, with local epochs as $10$. The batch size is $64$.

\subsubsection{Implementation of FedCLIP}

FedCLIP\cite{radford2021learning} adopts a pre-trained CLIP model to extract visual-textual joint embedding for multi-modal classification.

\paragraph{Architecture} The FedCLIP baseline mainly contains a pre-trained $vit\_b\_16$ as the image encoder and a pre-trained $BERT-base$ model as the text encoder. In addition, the model also has a projector head made of two linear layers mapping the output of the encoders into a latent space with $256$ as the embedding dimension. The concatenated embeddings will then be sent to a linear layer with ($512$, $num\_classes$).

\paragraph{Training}
The FedCLIP training is divided into two stages. The first stage is training the encoders and the projector head using contrastive loss proposed in \cite{radford2021learning}. The second stage is training with the supervised loss with the participation of the classifier layer. The server aggregates the uploaded models with FedAvg. The FedCLIP is trained with AdamW optimizer\cite{loshchilov2017decoupled}  with a learning rate of $2e-5$. The uploaded model will be aggregated by FedAvg.The total communication round is $200$, with local epochs as $10$. The batch size is $64$.

\subsubsection{Implementation of FedViLT}

FedViLT deploys ViLT proposed in \cite{ma2022multimodal} with multi-task training to improve the robustness towards the modality missing. 

\paragraph{Architecture}
FedViLT adopts standard ViLT model\cite{kim2021vilt} as feature extractor and follows with a linear layer as the classifier.

\paragraph{Training} In the local training process, we adopt the multi-task loss proposed in \cite{ma2022multimodal} to improve the model robustness to the modality missing. It shares the same hyperparameters as the above baseline models.

\subsubsection{Implementation of MMFed}
MMFed \cite{xiong2022unified} uses a co-attention module for local multi-modal fusion with meta-learning-based personalization. We use the model architecture and training schema as \cite{xiong2022unified} except for using pre-trained BERT to preprocess the text embedding and adopting the same training hyperparameters.

\subsubsection{Implementation of FedMVP}
\paragraph{Architecture}
The FedMVP adopts a pre-trained $vit\_b\_16$ as the image-specific encoder and a pre-trained $BERT-base$ model as the text-specific encoder. A joint encoder consists of two parallel cross-modal attention layers following a multi-head attention layer with $4$ attention heads. Finally, we employ a two-layer linear projection (proj1, proj2) and a classifier ($d_{latent}$, $num\_classes$) to categorize the output into logits. 

\paragraph{Training} 
Prior to the first communication round, the server-side pre-trained model will be broadcast to the clients. We adopt the linear warm-up scheduler ($10$ communication rounds), linear constant decay scheduler ($30$ communication rounds), and cosine annealing scheduler to control the learning rate. We provide more training details in Table~\ref{tab:fedmvp_para}.

\begin{table}[!htbp]
    \centering
    \caption{Training hyperparameters for FedMVP}
    \label{tab:fedmvp_para}
    \begin{tabular}{l l}
    \toprule
         \textbf{Hyperparameter} & \textbf{Value} \\ \midrule
           MCM loss scale $\alpha$ & $0.01$ \\
           RAM scale $\beta$ & $0.5$ \\
          MCM temperature $\tau$ & $0.1$ \\
            optimizer & AdamW \\
            initial learning rate & $2e-4$ \\
            batch size & $64$ \\
            communication round & $200$ \\
            local epochs & $10$ \\
            client aggregation ratio & $0.7$ \\
            random seed & $5481$ \\
            common dimension $d_{com}$ & $256$ \\
            latent dimension $d_{latent}$ & $512$ \\
    \bottomrule
    \end{tabular}
    
\end{table}

\section{Additional Numerical Experiments}

\begin{table}[!htb]
\centering
\caption{Experiments on CUB-200  dataset with different image text missing ratio under IID setting. Given $100$ image-text pairs, $0.5:0.5$ with $\beta=0.3$ represents among the $30$ incomplete image-text pair, $15$ image-text pairs lost their image and the rest lost their text data.}
\begin{tabular}{c ccc ccc}
\toprule
\multirow{2}{*}{\textbf{Image:Text Ratio}} & \multicolumn{3}{c}{\textbf{CUB-200}} & \multicolumn{3}{c}{\textbf{Oxford Flower}} \\ 
\cmidrule(r){2-7} 
& $\beta=0.3$ & $\beta=0.5$ & $\beta=0.8$ & $\beta=0.3$ & $\beta=0.5$ & $\beta=0.8$\\ 
\midrule
0.3:0.7 &    $78.12\%$      &    $73.01\%$      &    $71.06\%$  &    $93.67\%$      &    $92.10\%$      &    $90.15\%$   \\
0.5:0.5 &      $77.89\%$   &   $74.46\%$    &     $70.31\%$  &   ${93.19\%}$       &   ${91.28\%}$   &    ${89.32\%}$   \\ 
0.7:0.3 &     $77.06\%$     &    $72.68\%$      &    $68.85\%$   &     $92.79\%$     &    $90.42\%$      &    $87.94\%$  \\ \bottomrule
\end{tabular}
\label{tab:rt}
\end{table}

\textbf{Impact of Incomplete Modality.} In Table~\ref{tab:rt}, we show the impact of the proportion of image and text lost in modality missing on model performance. Notably, as the proportion of missing images increases, the model's performance slightly degrades, indicating that prompt-augmented image generation provides more training benefits than generated text data. As the overall modality missing rate increases, this trend becomes more pronounced. For instance, when $\beta$ increases from $0.3$ to $0.8$, the performance margin between $03:07$ Ratio and $07:03$ Ratio increases from $1.06\%$ to $2.21\%$. Similar observations can be obtained on the Oxford Flower dataset.

\section{Long-term Impact}
\label{sec:long_term}

The heterogeneity of data~\cite{kairouz2021advances,liang2021omnilytics} has consistently stood as one of the most significant challenges within the field of federated learning. Numerous existing studies have focused on addressing issues pertaining to class heterogeneity~\cite{10198520}, label heterogeneity~\cite{9671374,long2021fedsiam}, and model heterogeneity~\cite{wang2024towards},  yielding promising results. However, the modality heterogeneity~\cite{s23156986} of data has yet to garner widespread attention within the community. The multi-modal data~\cite{baltruvsaitis2018multimodal,10446126} in the real world is usually unpaired and stored in a distributed manner, which increases the difficulty in utilizing it for the training of advanced models. These unpaired or missing data will greatly degrade the model performance and hinder efficient global aggregation. On the other hand, the clients in a federated learning system usually have limited hardware resources and communication bandwidth\cite{10233897,NEURIPS2023_bc6a1f96,9992745,pmlr-v211-wang23d}, which makes it impractical to train from scratch or even fine-tune large-scale foundation models in FL scenarios~\cite{wang2023knowledge}. However, given the impressive performance and generalization capabilities of these pre-trained foundational models in multi-modal learning, the capability to integrate the large-scale pre-trained models in a lightweight way is expected for an effective federated multi-modal learning framework.  

To address the broadly existing modality incompleteness problem in distributed silos, we proposed a novel knowledge transfer-based multi-modal federated learning framework. Our proposed method FedMVP provides a solution for integrating large-scale pre-trained models to complete missing data and empower the federated system with resource limitations. This has practical significance for many real-world data science applications. For example, in collaborative efforts among multiple medical institutions, the patient examination and consultation process generates a large amount of multimodal data\cite{https://doi.org/10.1002/alz.068774,MA2024120609,Lyu2023-cd}, which often exists in the form of structured data, medical images, and text. However, differences in medical examination procedures and equipment often result in the absence and variability of different modalities of data. Utilizing the domain knowledge of foundation models can often capture more information. In the context of social media, there is significant application value in pairing text with images and captioning images with text when people publish information. We hope this work can provide some inspiration for future research in this field.
% \section{Broad Impact}

\bibliographystyle{splncs04}
\bibliography{ref}